\documentclass[a4paper, conference]{IEEEtran}

\usepackage{graphicx}      
\usepackage{amsmath,amsfonts}
\usepackage[scr=boondox]{mathalpha}
\usepackage{array}
\usepackage[caption=false,font=normalsize,labelfont=sf,textfont=sf]{subfig}
\usepackage{stfloats}
\usepackage{url}
\usepackage{verbatim}
\graphicspath{{Graphics/}}
\usepackage{layouts}
\usepackage{siunitx}
\usepackage{bm}
\usepackage[acronym]{glossaries}
\usepackage[]{todonotes}
\usepackage{textcomp}
\usepackage[ruled,vlined,algo2e]{algorithm2e}
\usepackage{authblk}
\usepackage{layout}
\makeatletter
\def\@makefntext#1{%
  \setlength{\parindent}{0pt}%
  \leavevmode\@makefnmark#1%
}
\makeatother

\newacronym{mpm}{MPM}{Material Point Method}
\newacronym[]{fem}{FEM}{Finite Element Method}
\newacronym{mppi}{MPPI}{Model Predictive Path Integral}
\IEEEoverridecommandlockouts
\begin{document}
\title{Differentiable Material Point Method \\
for the Control of Deformable Objects} 

\author[1,2]{Diego Bolliger}
\author[1]{Gabriele Fadini}
\author[2]{Markus Bambach}
\author[1]{Alisa Rupenyan%
\thanks{This research was partially supported by Rieter AG and the Johann Jacob Rieter Foundation, and by NCCR Automation, a National Center of Competence in Research, funded by the Swiss National Science Foundation (grant number 51NF40\_225155).}%
\thanks{\texttt{\{boie,fadi,rupn\}@zhaw.ch}, \texttt{mbambach@ethz.ch}}}

\affil[1]{\textit{Centre for Artificial Intelligence, ZHAW}}
\affil[2]{\textit{Advanced Manufacturing Laboratory, ETH Zürich}}

\maketitle

\begin{abstract}
    Controlling the deformation of flexible objects is challenging due to their non-linear dynamics and high-dimensional configuration space. This work presents a differentiable Material Point Method (MPM) simulator targeted at control applications. We exploit the differentiability of the simulator to optimize a control trajectory in an active damping problem for a hyperelastic rope. 
    The simulator effectively minimizes the kinetic energy of the rope around $2\times$ faster than a baseline MPPI method and to a 20\,\% lower energy level, while using about $3\,\unit{\percent}$ of the computation time.
\end{abstract}



\section{Introduction}
Controlling the deformation of flexible objects is a challenging task. Their behavior is governed by nonlinear partial differential equations and an effectively infinite-dimensional state and configuration space. 
This limits the applicability of traditional control methods, particularly for large, non-linear deformations or bodies with varied properties and materials.

Various approaches have been explored for controlling the deformation of a flexible object. 
Although general analytical solutions are not possible, in simple cases, control laws can be found by discretizing the object into finite elements.
This is demonstrated by \cite{fonkouaDeformationControl3D2024}, where locally valid Jacobians are evaluated and used for inversion-based control. Further, \cite{mylapilliControlThreedimensionalIncompressible2017} find nonlinear control laws by analytically solving the cantilever beam equation and recasting control objectives as constraints.
The control laws obtained by these methods are at least locally guaranteed to drive the body towards the desired configuration, and they are efficient to implement. However, generalizing them to problems with complex shapes or material models is difficult and requires ad-hoc derivation for every case.

Reinforcement learning has also been applied to deformation control in \cite{hanModelbasedReinforcementLearning2017, liangRealtoSimDeformableObject2024}. These methods require extensive training data, which is costly to collect.
As a more data-efficient approach, learning from demonstrations was explored in papers such as \cite{balaguerCombiningImitationReinforcement2011, leeLearningForcebasedManipulation2015}, but this leads to a control policy that is restricted by the used demonstration and less generalizable. For higher-level action planning, \cite{sundaresanLearningRopeManipulation2020, zhouLaSeSOMLatentSemantic2021} embed the state of an object into a learned latent space, where planning is easier, and then transform the actions back into actual controls. 
This method enables moving an object between static configurations, whereas dynamic object or goal behavior is not incorporated.

Recently, differentiable simulations have been explored, recovering simulation gradients and offering new opportunities for material identification and control.
For example, \cite{liuRoboticManipulationDeformable2023, chenDifferentiableDiscreteElastic} implemented differentiable position-based dynamics (PBD) and discrete elastic rod (DER) simulations for controlling rope or cable shapes. While they provide promising directions, PBD is an algorithm stemming from computer graphics and animation, which may limit physical realism, and DER is restricted to one-dimensional objects, which prevents generalization. \cite{jatavallabhulaSimDIFFERENTIABLESIMULATION2021} propose a differentiable simulator and rendering engine for soft bodies using \acrfull{fem}, and \cite{huChainQueenRealTimeDifferentiable2019, huangPlasticineLabSoftBodyManipulation2021} introduce differentiable \acrshort{mpm} simulators, with the latter providing a benchmark collection of manipulation tasks. Additionally, \cite{zhengDifferentiableClothParameter2024} use a differentiable simulator to learn cloth parameters. These works demonstrate the use of differentiable simulators for controlling or manipulating bodies such as a soft-body walker, cloth, or clay. Still, the use of these simulators in dynamic control problems according to a time-varying objective rather than a static configuration, has not been fully analyzed and demonstrated. Furthermore, the \acrshort{mpm} variants in \cite{huangPlasticineLabSoftBodyManipulation2021, huChainQueenRealTimeDifferentiable2019, liuRoboticManipulationDeformable2023} are designed for computer graphics and prioritize performance at the cost of accuracy, while the use of FEM in \cite{jatavallabhulaSimDIFFERENTIABLESIMULATION2021} is precise but computationally expensive and can struggle with large deformations.
In this work, we propose a  differentiable, \acrfull{mpm}-based simulator targeted at control and apply it in trajectory optimization for a dynamic control task.
By focusing on dynamic control and adapting the simulator for control, our approach provides a novel perspective on deformable object manipulation.
The main contributions of the paper are as follows:
\begin{enumerate}
    \item We present a simulator, based on \acrshort{mpm}, tailored for optimal control involving deformable objects.
    \item We build on standard \acrshort{mpm} implementation, which we modify such that it is differentiable, can be accelerated with GPUs, and  accurately models the system's mechanical energy.
    \item We showcase a control approach using the proposed simulator for an active damping scenario, a dynamic problem not previously explored and compare its performance to a baseline sampling-based control method.
\end{enumerate}

\section{Problem Formulation}

Our goal is to control the deformation of a deformable body $\mathcal{B}$ using optimal control. 
To model $\mathcal{B}$ we choose the continuum mechanics approach, where $\bm{X}$ represents the infinitesimal points that constitute the domain of $\mathcal{B}$.
The motion of the deformable body is then described through a mapping $\bm{x} = \phi(\bm{X}, t)$ where $\bm{x}$ are the positions of the points of $\mathcal{B}$
at any time $t$. This mapping, and thus the body's motion, is governed by a system of non-linear partial differential equations. The full formulation can be found in \cite{nguyenMaterialPointMethod2023}. For our purposes, the main governing terms are:
\begin{align*}
	\frac{D \rho}{Dt}+\rho\nabla \cdot \bm{v} = 0, && \text{conservation of mass} \\
	\rho \frac{D \bm{v}}{Dt} = \nabla \boldsymbol{\sigma} + \rho \bm{b}, && \text{conservation of linear momentum}
\end{align*}
where $\rho(\bm{X},t)$ is the density, $\bm{v}(\bm{X},t)$ the velocity, $\bm{b}(\bm{X},t)$ are body forces (e.g. gravity and external forces), and $\bm{\sigma}(\bm{X},t)$ is the Cauchy stress, which we obtain through a material model. Furthermore, the body is subject to initial conditions $\bm{v}(\bm{X},0)=\bm{v}_0$, $\bm{\sigma}(\bm{X}, 0) = \bm{\sigma}_0$, and to boundary conditions, such as $\bm{v}(\bm{X},t)=\bm{\Bar{v}}$, on the body's boundary $\partial \mathcal{B}$.
For completeness, we also introduce the deformation gradient $F = \frac{\partial \bm{x}}{\partial \bm{X}}$ which measures the deformation of an infinitesimal element of $\mathcal{B}$.

Control on this body can now be exerted through body forces $\bm{b}(\bm{X},t)$ or through the boundary conditions. However, this is a continuous-time and infinite-dimensional problem that cannot be generally solved in closed form. Hence, we simulate the object with an \acrshort{mpm} where the body is discretized into a finite number of particles and converted into a discrete-time problem. 

\subsection{Optimal Control Problem}
Using \acrshort{mpm} the body $\mathcal{B}$ is described as a set of $P$ particles $p$, each representing a finite volume $V_p$ of the body. This leads to the particle states
\begin{equation}
    \bm{S}_p^t =  \left[\bm{x}_p^t, \bm{v}_p^t, \bm{F}_p^t\right]^T
\end{equation}
where $\bm{x}_p^t$ is the particle's location, $\bm{v}_p^t$ is the particle's velocity and $\bm{F}_p^t$ is its deformation gradient at a time $t$. The full state of the body then is $\bm{S}^t = [\bm{S}_1^t; \dots; \bm{S}_P^t ]$ and we can express the evolution of the discrete-time system as
\begin{equation}
\label{eq:discrete_sys}
    \bm{S}^{t+1} = \mathscr{f} (\bm{S}^t, \bm{u}^t, t).
\end{equation}
Here, the control $\bm{u}^t$ also can either enter the dynamics through the discretized body forces $\bm{b}_p^t$ (e.g. external forces) or be enforced as part of the boundary conditions (e.g. imposed motion);
in both views, the control is simply a parameter of the simulation function $\mathscr{f}$.
Finally, this allows us to formulate the discrete-time optimal control problem
\begin{align}
\label{eq:opt_problem}
    \bm{u^*} = \arg &\min_{\bm{u}} \sum_{t=0}^{T} c^t(\bm{S}^t, \bm{u}^t) \\
    \text{subject to: } & \bm{S}^{t+1} = \mathscr{f}(\bm{S}^t, \bm{u}^t) \nonumber \\
    & \text{initial \& boundary conditions} \nonumber
\end{align}
where $c^t$ is a time-varying, state-dependent cost and $\bm{u} = \{\bm{u}^0, \bm{u}^1, \dots, \bm{u}^{T-1}\}$ the discrete-time inputs. 
In the following, we address how we discretize and simulate the continuum body, and how we can then solve this resulting optimal control problem to control the object's deformation.
\section{Background}
\label{sec:background}

\subsection{\acrfull{mpm}}
\label{ssec:flip_mpm}
We use the Material Point Method (\acrshort{mpm}) to numerically solve the continuum body problem. Unlike \acrshort{fem}, \acrshort{mpm} avoids meshing or re-meshing, and unlike fully mesh-free methods, it does not require neighborhood searches. This method has been applied in fields such as geotechnical engineering and computer graphics.

Using the standard Fluid Implicit Particle (FLIP) update-stress-last (USL)  \acrshort{mpm} described in \cite{nguyenMaterialPointMethod2023},
the material is discretized into particles $p$, each storing position $\bm{x}_p$, velocity $\bm{v}_p$, mass $m_p$, volume $V_p$, deformation gradient $\bm{F}_p$ and stress $\bm{\sigma}_p$ at each time step.
These particles interact with a fixed Eulerian grid, whose nodes $I$ temporarily store grid masses $m_I^t$, momenta $m\bm{v}_I^t$, forces $\bm{f}_I^t$ and velocities $\bm{v}_I^t$. The grid nodes reset to zero after every time step.
At each time step $t$ in the simulation horizon $T$, the method performs the following steps:

\begin{enumerate}
	\item Particle to Grid (P2G) step: Particle masses $m_p^t$ and momenta $m_p^t \bm{v}_p^t $ are transferred to the grid nodes, respecting mass and momentum conservation.
	Further, nodal forces $\bm{f}_I^t$ are computed from particle stresses $\bm{\sigma}_p^t$ and body forces $\bm{b}_p^t$.
	
	\item Grid operations: Grid momenta are updated by taking a single Euler integration step:\\
    ${m  \bm{v}_I^{t+1} = m \bm{v}_I^t+\bm{f}_I^t \Delta t}$.\\
	Then nodal velocities $\bm{v}_I^t$ and $\bm{v}_I^{t+1}$ are found by dividing momenta by mass.
    \label{item:step_2}
	\item Grid to particle (G2P) step: In a second Euler integration step, the new particle location $\bm{x}_p^{t+1}$ and $\bm{F}_p^{t+1}$ are found from the new nodal velocities $\bm{v}_I^{t+1}$. Further, the old and new nodal velocities  $\bm{v}_I^{t}$ and $\bm{v}_I^{t+1}$ are used to update particle velocities $\bm{v}_p^{t+1}$. Finally, the volumes $V_p^{t+1}$ are evaluated, and the particle stresses $\bm{\sigma}_p^{t+1}$ are found through the material model (section \ref{sec:num_exp}).
    \label{item:step_3}
\end{enumerate}
The control action $\bm{u}^t$ is either injected into the simulation with the body forces or enforced as part of the boundary conditions.

\subsection{Control Optimization with Automatic Differentiation}
Given the high number of degrees of freedom, the optimization presented in \eqref{eq:opt_problem} is rather challenging. 
To this end, we use gradient-based methods, which can find locally optimal solutions and require only the computation of first-order derivatives of the cost function for any optimization step.
However, the computation of analytical gradients for a full \acrshort{mpm} is prohibitively complex  and not flexible if the setting changes.
To cope with this issue, several computational frameworks have been developed to compute the gradients algorithmically through automatic differentiation (AD).
In practice, AD leverages the fact that any function can be expressed in terms of basic operations and calculates the gradients by automatically composing the derivatives of basic operations in the computation graph using the chain rule.
In particular, machine learning libraries, such as PyTorch, TensorFlow, {Taichi-Lang}, or JAX, implement AD and GPU parallelization of Python code.
Other frameworks also exist for C++, Julia, and other languages through projects like Clad or Enzyme.
Among these AD alternatives, we decided to use JAX \cite{jax2018github} because it is mature, under active development, and can be easily integrated with other machine learning libraries.
Additionally, we use Adam \cite{Kingma2014AdamAM}, an advanced gradient-based optimizer with adaptive learning rates, to achieve a more robust optimization once the gradients are computed.
Finally, one of the main issues of continuum mechanics-based simulations, such as FEM or MPM, is that they are computationally expensive.
To speed up their computation, we parallelized particle and grid operations.
The capability of JAX to interface with GPUs proved to be fundamental to this end, allowing us to write code that can be automatically ported and parallelized.

\subsection{Model Predictive Path Integral}
\label{sec:mppi}
To our best knowledge, optimal control with deformable objects in the setting that we aim to tackle does not have an analytical solution.
As a baseline, we compare our gradient-based method with \acrfull{mppi} \cite{williamsModelPredictivePath2017}, a sampling-based reactive control method. We choose it because it is frequently used in robotics and offers an alternative way to tackle optimal control problems such as \eqref{eq:opt_problem}
without using gradients. It does so by selecting the control action that achieved the lowest cost, by simulating a bundle of control trajectories.
Assuming discrete-time stochastic dynamics of the form 
\begin{align}
	\bm{x}^{t+1} = \mathscr{f}(\bm{x}^t, \Tilde{\bm{u}}^t),  &&  \Tilde{\bm{u}}^t \sim \mathcal{N}(\bm{u}^t, \bm{\Sigma}).
\end{align}
MPPI samples a set of noisy control input sequences $\bm{\Tilde{u}_k} = \{\Tilde{\bm{u}}_k^0, \dots, \Tilde{\bm{u}}_k^{T-1}\}$ and simulates $K$ trajectories $\bm{q_k} = \{\bm{x}^0, \mathscr{f}(\bm{x}^0, \Tilde{\bm{u}}_k^0), \dots, \mathscr{f}(\bm{x}_k^{T-1}, \Tilde{\bm{u}}_k^{T-1})\}.$
Each trajectory incurs a total cost $C_k = \sum_{t=0}^T c_k^t(\bm{x}_k^t, \bm{u}_k^t)$. The optimal control sequence is then obtained as a weighted sum of the noisy input sequences
\begin{align}
	\bm{u}^\star = \sum_{k=1}^{K} w_k \bm{\Tilde{u}}_k
    \label{eq:mppi_eq}
\end{align}
where
\begin{align}
	w_k = \frac{1}{\eta} \mathrm{exp}\left(-\frac{1}{\beta} (C_k - \rho)\right), && \sum_{k=1}^{K} w_k = 1.
\end{align}

Here, $\beta$ is the inverse temperature and $\eta$ is the normalization constant to achieve partition of unity.
To achieve closed-loop control, this scheme is then applied in a receding horizon approach. At each timestep $t$ we simulate trajectories of length $H < T$ rather than the full trajectory, starting from the current state $\bm{x}^t$.  Then only the first input of the resulting input vector ${\bm{u}_H^t}^\star$ is used as the control input at $t$ and at $t+1$ a new optimal control sequence ${\bm{u}_H^{t+1}}^\star$ is evaluated. 
Following \cite{pezzatoSamplingBasedModelPredictive2025}, we further conduct a line search each time step to assure that the normalization constant is in a range $\eta \in (\eta_{min}, \eta_{max})$.
\section{Proposed Method}
In this section, we present an overview of our differentiable \acrshort{mpm} simulation, including the algorithm and implementation details.
\subsection{Control-Oriented Material Point Method}
We adapt the FLIP \acrshort{mpm} algorithm from section \ref{ssec:flip_mpm} to tailor it for control optimization.
At low spatial resolution (coarse grid and small number of particles), this standard method suffers from strong numerical dissipation, meaning it does not properly conserve energy but instead dissipates it through numerical errors. This leads to an erroneous damping of the system, which is problematic for a control application where we need to accurately simulate energy to capture stability properties. Only a very high-resolution simulation then can be used for optimizing control inputs with a standard \acrshort{mpm}, leading to prohibitively long computational times. Instead, we aim at simulating the system using a coarser spatial resolution, while maintaining energy conservation.
Numerical dissipation is a known problem in \acrshort{mpm} simulations, and a discussion can be found in \cite{nguyenMaterialPointMethod2023}. The dissipative behavior depends on different factors, such as the procedure used to transfer quantities between the grid and particles or whether the stress is updated at the beginning or end of the simulation step. 

We found that the numerical dissipation can be significantly improved compared to FLIP MPM by changing the integration procedure. In section \ref{ssec:flip_mpm} we see that there are two Euler integration steps. First, in step \eqref{item:step_2}, the grid velocities $\bm{v}_I^{t+1}$ are integrated. Afterward, in step \eqref{item:step_3}, the new grid velocities are used to integrate the location and deformation gradient of the particles.
In contrast, we rearrange the algorithm so that it only calculates the derivatives of the particle state $\bm{S}_p^t =  \left[\bm{x}_p^t, \bm{v}_p^t, \bm{F}_p^t\right]^T$. This allows us to express the system as a first-order ordinary differential equation with
\begin{equation}
\label{eq:derivative}
   \dot{S}^{t}= \Tilde{\mathscr{f}}(\bm{S}^t, t),
\end{equation}
and integrating it with any explicit or implicit integration solver to advance the system. This formulation also provides a canonical form for applying traditional control methods.

A reduced version of the algorithm is shown in Algorithm \ref{alg:alg1}.
To obtain the derivatives of the particle quantities, the following steps are taken:

\begin{enumerate}
	\item P2G: Obtain node mass $m_I^t$ and momenta $m_I^t \bm{v}_I^t$.
	\item G2P2G: Interlaced grid-to-particle-to-grid step; Using node momenta, find derivative of deformation gradient $\dot{\bm{F}}_p^t$ and position $\dot{\bm{x}}_p^t$. Then calculate stresses $\bm{\sigma}_p^t$. Finally, find node forces $\bm{f}_I$ from stresses and body forces $\bm{b}_p^t$.
	\item Grid operations: Calculate node accelerations $\bm{a}_I$.
	\item G2P: Use node accelerations to find derivative of velocities $\dot{\bm{v}}_p^t$.
\end{enumerate}

Evolving the simulation can now be performed as usual, for example, using Euler integration
\begin{align}
	\bm{S}^{t+1} = \bm{S}^t + \Delta t \Tilde{\mathscr{f}}(\bm{S}^t, t)
\end{align}
or, more generally, by using any explicit integration scheme $G$ as
\begin{align}
	\bm{S}^{t+1} = \bm{S}^t +  G\left(\Tilde{\mathscr{f}}(\bm{S}^t, t)\right).
\end{align}

For our purposes, we use Runge-Kutta-4 (RK4) to integrate the system, which leads to a simulation with negligible numerical dissipation. 
The control $\bm{u}^t$ is then again injected through body forces or enforced as boundary conditions during integration, leading to a system of form \eqref{eq:discrete_sys}.

\begin{algorithm2e}[h!]
    \small
	\SetAlgoVlined
	\DontPrintSemicolon
	\KwIn{$\bm{x}_p, \bm{v}_p, V_p, \bm{F}_p, m_p, \bm{b}_p$}
	\KwOut{$\dot{\bm{x}}_p, \dot{\bm{v}}_p, \dot{\bm{F}}_p$}
	\SetKwFunction{ptg}{P2G}
	\SetKwFunction{gtpstar}{G2P*}
	\SetKwFunction{gtp}{G2P}
	\SetKwProg{gtptg}{G2P2G}{:}{}
	 
	$m_I, \bm{v}_I \leftarrow$ \ptg{$m_p, \bm{v}_p$}\;
	\tcp{Transfer particle mass and velocity to the grid.}
	\gtptg{($\bm{v}_p, V_0, \bm{F}_p, m_p, \bm{v}_I, \bm{b}_p$)}{
		\textbf{Evaluate} $\dot{\bm{x}}_p, \dot{\bm{F}}_p$ \textbf{from} $\bm{v}_I$\;
		\tcp{Derivatives of particle location \& deformation gradient from grid velocity.}
		\textbf{Evaluate} $V_p, \bm{\epsilon}_p, \dot{\bm{\epsilon}}_p$\ \textbf{from} $\bm{F}_p, \dot{\bm{F}}_p$\;
		\tcp{Particle volume, strain, strain-rate, Eq.\eqref{eq:green_strain},\eqref{eq:diss_strain}.}
		\textbf{Evaluate} $\bm{f}_I$ \textbf{from} $\bm{x}_p, V_p, \bm{\epsilon}_p, \dot{\bm{\epsilon}}_p$, $\bm{b}_p$\;
		\tcp{Internal forces from particle data, Eq. \eqref{eq:cauchy},\eqref{eq:diffmodel}.}
		\Return $\dot{\bm{x}}_p, \dot{\bm{F}}_p, \bm{f}_I$\;
	}
	$\bm{a}_I \gets \bm{f}_I / m_I$\tcp*[f]{Grid Acceleration}\;
	$\dot{\bm{v}}_p \gets$ \gtp{$\bm{v}_I, \bm{a}_I$}\;
	\tcp{Particle acceleration from grid.}
	\Return $\dot{\bm{x}}_p, \dot{\bm{v}}_p, \dot{\bm{F}}_p$\;
 	\caption{Particle Derivative Evaluation, Eq. \eqref{eq:derivative}}
	\label{alg:alg1}
\end{algorithm2e}

\subsection{Differentiable Implementation}
The simulation is implemented such that it can collect a running cost during execution and differentiate this cost with respect to various quantities
governing the simulation. 
Technically, this optimization variable could be any of the quantities required for the simulation, such as material properties, body forces or boundary conditions. 

\begin{figure}[t]
    \includegraphics[width=\columnwidth]{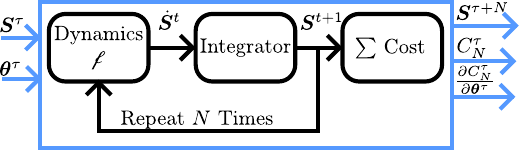}
    \caption{Inner loop of hierarchical simulation architecture. Gradients can be calculated through N time steps.}
    \label{fig:DiffArch}
    \centering
\end{figure}

Our method splits the entire simulation rollout into smaller batches of length $N$ due to the long horizon.
This results in a nested structure in which an inner simulation loop advances the simulation by $N$ steps and an outer loop collects the gradients
and optimizes the parameters, or controls, governing the simulation.
Such an outer loop repeatedly calls the inner loop until a full trajectory is obtained.
This approach is illustrated in Figure~\ref{fig:DiffArch}. The $N$-step inner function is compiled, runs on a GPU, and can be differentiated.
When calling the function, we supply the state $\bm{S}^{\tau}$, auxiliary data such as grid size, and the optimization variable $\bm{\theta}^\tau$.

At each time step $t = \tau, \tau+1, ... ,\tau+N$, the N-step-simulation function utilizes algorithm \ref{alg:alg1} to obtain the time derivatives $\dot{\bm{S}^t}$ of the state and integrates the system to obtain $\bm{S}^{t+1}$. In addition, we collect the partial cost
\begin{equation}
	C_N^\tau = \sum_{t=\tau+1}^{\tau+N} c^t(\bm{S}^t, \bm{u}^t).
	\label{eq:state_cost}
\end{equation}
After $N$ steps, the function returns the updated state $\bm{S}^{\tau+N}$ and the entire collected cost $C_N^\tau$.
Using AD, we can additionally obtain the derivative of the cost with respect to the optimization variables $\bm{\theta}^\tau$
\begin{equation}
	\frac{\partial C_N^\tau}{\partial \bm{\theta}^\tau}.
\end{equation}
The horizon length $N$ of the inner loop determines how much of the dynamic effects of $\bm{\theta}$ we capture.
Assuming $\bm{\theta}^\tau = \bm{u}^\tau$ is the control input at $t = \tau$, we have to assume that $\bm{\theta}^\tau$ also affects all states $\bm{S}^{\tau +N+1}, \dots, \bm{S}^T$ and the cost \eqref{eq:state_cost} might not adequately capture the effect of $\bm{\theta}^\tau$. Hence, when selecting $N$, we have to balance accuracy and computational cost.

\section{Numerical Experiments}
\label{sec:num_exp}
In this section, we describe the numerical experiments conducted to test our method for control problems.

We model a hyperelastic material using the Saint Venant-Kirchhoff model, to demonstrate the proposed approach. This is a memoryless constitutive model where the stress depends entirely on the current strain. Using the Green strain tensor
\newcommand{\Tr}[0]{\mathrm{Tr}}
\newcommand{\I}[0]{\mathrm{I}}
\begin{equation}
\label{eq:green_strain}
     \bm{\epsilon}_p^t = \frac{1}{2} ({\bm{F}_p^t}^T\bm{F}_p^t - \bm{\I}),
\end{equation}
we calculate the Cauchy stress as
\begin{align}
\label{eq:cauchy}
	\bm{\sigma}_{s,p}^t &= \bm{F}_p^t \bm{P}_{s,p}^t {\bm{F}_p^t}^T \quad \text{with} \\ \nonumber
    \bm{P}_{s,p}^t & = \lambda \Tr(\bm{\epsilon}_p^t) \bm{\I} +2\mu\bm{\epsilon}_p^t,
\end{align}
where $\lambda$ and $\mu$ are Lamé parameters and $\bm{P}_{s,p}^t$ is the Piola-Kirchhoff stress.
The hyperelastic model on its own does not contain any damping. To represent more realistic behavior, we utilize the strain-rate dependent damping model presented in \cite{m.sanchez-banderasStrainRateDissipation2018}. This leads to a damping term that is analogous to the stress, but with respect to the strain rate. Using the time derivative of the strain
\begin{align}
\label{eq:diss_strain}
	\dot{\bm{\epsilon}}_p^t = \frac{1}{2} \left({\bm{F}_p^t}^T \dot{\bm{F}_p^t}+ \dot{{\bm{F}_p^t}}^T \bm{F}_p^t\right) ,
\end{align}
we find the dissipation stress
\begin{align}
	\bm{\sigma}_{d,p}^t & = \bm{F}_p^t \bm{P}_{d,p}^t  {\bm{F}_p^t}^T  \quad \text{with} 
	\label{eq:diffmodel}\\ \nonumber
    \bm{P}_{d,p}^t & = \lambda_d \Tr(\dot{\bm{\epsilon}}_p^t) \bm{\I} +2\mu_d \dot{\bm{\epsilon}}_p^t,
\end{align}
where $\lambda_d$, $\mu_d$ are damping parameters.
The final stress then results to
\begin{align}
	\bm{\sigma}_p^t =\bm{\sigma}_{s,p}^t + \bm{\sigma}_{d,p}^t.
\end{align}

\subsection{Energy Conservation}
To verify the method, we consider the conservation of mechanical energy in simulation. For this, we simulate a flexing beam in an isolated environment where it experiences no external forces.  We set up a hyperelastic beam, choosing the material properties to correspond to silicon rubber.
The beam is modeled to be hyperelastic, and the velocities of its particles are initialized with
\begin{align}
    v_{p,x} = 0, &&
    v_{p,y} =A \cos\left(2\pi\frac{x_{p,x}-x_{0,x}}{h_x}\right) ,
\end{align}
where $A$ is the velocity amplitude and $h_x$ is the length of the beam.
This velocity profile will elicit a flexing motion in the beam where both edges flex up and down relative to the center, and the beam as a whole moves upwards slowly.
We record the kinetic, strain, and total energies in the system to analyze numerical dissipation of the simulation.

\begin{figure}
    \centering
    \includegraphics[width=\columnwidth]{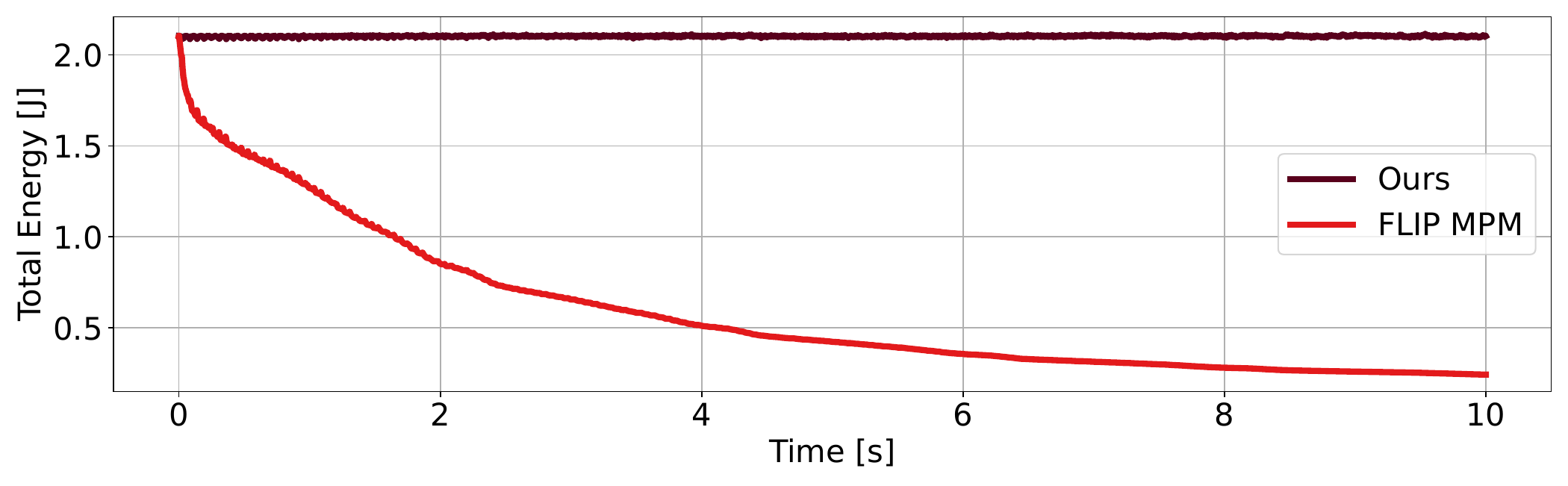}
    \caption{Our method (dark red) conserves the total energy (sum of strain and kinetic)
    while standard FLIP MPM (light red) quickly dissipates it.
    }
    \label{fig:flex_energy}
\end{figure}

Figure \ref{fig:flex_energy} compares the total energy trajectories of the flexing beam for our \acrshort{mpm} implementation and the standard \acrshort{mpm}. To make the comparison more representative, the standard method has a four times smaller integration step than our method, which uses RK4. Since the beam is simulated without friction and no gravity, the total energy of the system should remain constant at all times, and any energy loss stems from numerical errors.
Our implementation fulfills this requirement, retaining the total energy in the system for the entire simulation duration. The same behavior can also be observed over much longer time horizons.
The standard \acrshort{mpm} however dissipates energy very quickly, resulting in a loss of $88\,\%$ of the total energy after $10\, \unit{\s}$. Hence, the proposed control-oriented differentiable simulator is better suited for simulating low resolution problems over long time horizons.

\subsection{Trajectory Optimization}
\label{sec:traj_opt}
\begin{figure}[]
    \centering
    \includegraphics[width=\columnwidth]{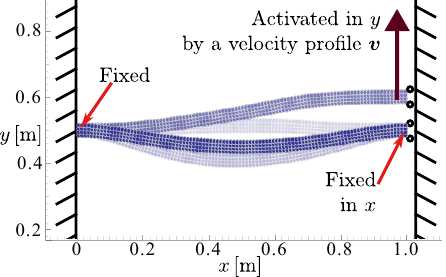}
    \caption{Simulation Experiment: Hyperelastic rope, fixed on the left and fixed in x direction on the right. The vertical velocity of the right rope end can be controlled. The simulation starts from a steady state where the rope is hanging through, and at first, it is excited to give rise to oscillations.}
    \label{fig:exp_setup}
\end{figure}

We use the differentiable simulator to optimize a velocity trajectory for active damping of oscillations. For this, we simulate a 2D \qtyproduct{1 x 0.04}{\meter} hyperelastic rope under the influence of gravity. The material properties resemble silicon rubber with a Young's modulus of $E = 1.5 \unit{\MPa}$ a Poisson ratio of $\lambda = 0.47$ and an initial density of $\rho_p = 1100 \unit{\kg\per\meter\cubed}$.
The left end of the rope is fixed, and the right end of the rope is fixed in the $x$ direction but can be controlled in the $y$ direction with a prescribed velocity $v_y^t$. This means control enters the simulation as a Dirichlet boundary condition. 
Starting from a hanging steady state, we first apply a scheduled velocity sequence $\bm{v^\text{init}}$ to bring kinetic energy into the system and elicit oscillations, then our goal is to find a velocity trajectory $\bm{u} = \left \{v_y^0, v_y^1, \dots , v_y^{T} \right \}$ that minimizes the kinetic energy, effectively damping oscillations. Figure \ref{fig:exp_setup} shows this simulation setup.

Formally, we define the state cost
\begin{equation}
    c^t(\bm{S}^t, \bm{u}^t) = \frac{1}{2} m_p {\bm{v}_p^t}^T \bm{v}_p^t.
    \label{eq:kin_cost}
\end{equation}

\begin{figure}
    \centering
    \includegraphics[width=\columnwidth]{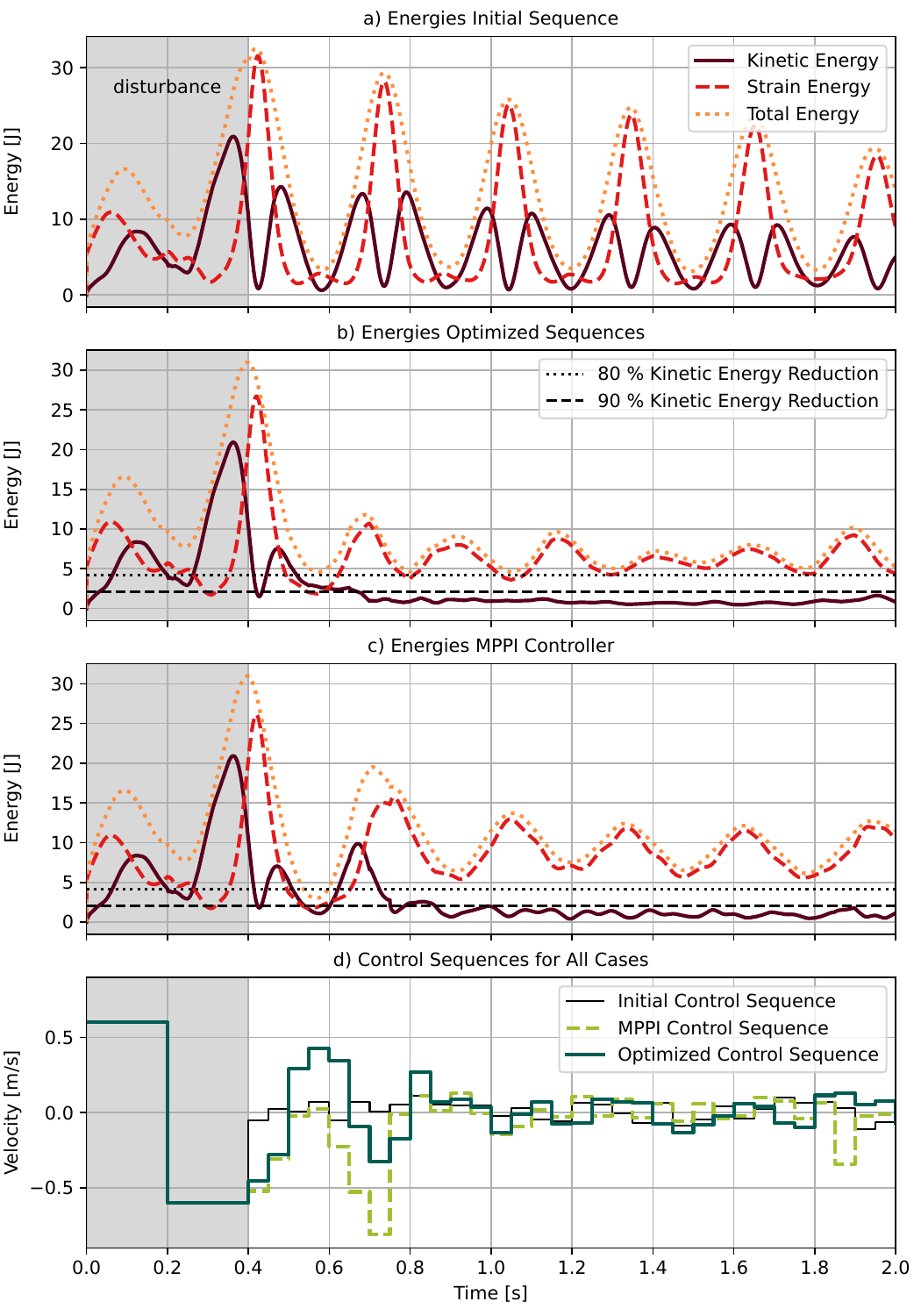}
    \caption{Energy trajectories and control sequences: a) Energies for the initial, random control sequence, b) Energies for the control sequence after optimization using the diff. simulator, c) Energies for the \acrshort{mppi} controller, d) Comparison of control sequences.}
    \label{fig:combined_plot}
\end{figure}

To solve problem \ref{eq:opt_problem} we now use gradient-based optimization, where we first select a candidate sequence $\bm{u}_0$ and then apply the following steps until convergence or for a set number of times:

\begin{enumerate}
    \item Full system simulation using $\bm{u}_n$, giving rise to the cost $C_n$.
    \item Differentiation of the cost with respect to control sequence 
    \begin{equation*}
        \nabla_{\bm{u}_n} C_n.
    \end{equation*}
    \item Optimization step using a gradient method, such as gradient descent
    \begin{equation*}
        \bm{u}_{n+1} = \bm{u}_n - \eta \nabla_{\bm{u}_n}C_n.
    \end{equation*}
    
\end{enumerate}
The trajectory optimization problem has a total duration of $2\,\unit{\s}$. In the first $0.4\,\unit{\s}$, a disturbance is applied to the system in the form of a fixed control sequence, and in the remaining $1.6 \, \unit{\s}$ we optimize the control sequence. To limit the dimensionality of the control sequence, we discretize it using zero-order-hold, where each value is held for $50 \, \unit{\ms}$, resulting in a total of $32$ applied velocities. Initially, all velocities are drawn uniformly from the interval $[-0.1, 0.1] \, \unit{\meter\per\second}$. According to the differentiable implementation, we further run the optimization in $0.4 \, \unit{\s}$ intervals, i.e., with $N=8$. Thus, the velocities applied in the interval $[0.4, 0.8] \, \unit{\s}$ are optimized independently of all other velocities, and the same is true for all following $0.4 \, \unit{\s}$ intervals.
We optimize the trajectory using Adam for a total of $50$ optimization steps.
Figure \ref{fig:combined_plot}.d) shows the initial and optimized velocity sequences, and  \ref{fig:combined_plot}.a) and \ref{fig:combined_plot}.b) show the resulting energy profiles. The optimization leads to a velocity profile with a visible oscillatory behavior that effectively damps the system’s oscillations and keeps the kinetic energy close to zero.

We compare our method with an MPPI controller, presented in Section \ref{sec:mppi}, as a baseline. For this example, we use a horizon of $H=10$ control steps, diagonal variance matrix $\bm{\Sigma} = 0.4 \I$ and utilize the mean $\bm\mu_t = \{\bm{u}^\star_{t-1, 1}, \dots, \bm{u}^\star_{t-1, H-1}, 0\}$ and $\bm\mu_0 = \bm{0}$. At every control step, 2000 trajectories are simulated, and, using the same state cost \eqref{eq:kin_cost} as for the gradient-based optimization, we find an approximation to the optimal control input according to equation \eqref{eq:mppi_eq}.
The resulting energy trajectory and input sequence are depicted in figures \ref{fig:combined_plot}.c) and \ref{fig:combined_plot}.d). The MPPI significantly damps oscillations in the system but causes visible noise due to its stochastic approach. In contrast, our method achieves damping faster and leads to a smoother trajectory because of the gradient-based approach.

As numerical metrics, we examine the time it takes both control methods to damp the system by 80\unit{\percent} and 90\unit{\percent} of the maximal kinetic energy caused by the disturbance. These two energy levels are indicated in figure \ref{fig:combined_plot}.a) and \ref{fig:combined_plot}.b) and we measure the time between the end of the disturbance at $0.4\,\unit{\s}$ and the last time at which the kinetic energy crosses the line and stays below it.  Further, we compare the mean kinetic energy in the $[1, 2] \, \unit{\s}$ interval. The results are listed in Table \ref{tb:kpi}. The maximal kinetic energy introduced by the disturbance is $20.9\,\unit{\J}$ and hence a $80\,\unit{\percent}$ reduction corresponds to $4.19\,\unit{\J}$, and a $90\,\unit{\percent}$ reduction to $2.09\,\unit{\J}$. As a reference, we include the mean energy of the kinetic energy for the undamped system, denoted by \textit{No Action}. Compared to \acrshort{mppi}, direct optimization achieves damping by $80\,\unit{\percent}$ $2.7 \times$ faster and damping by $90\,\unit{\percent}$ $1.7 \times$ faster. The mean energy in the interval $[1, 2] \, \unit{\s}$ is $20\,\unit{\percent}$ lower for direct optimization. The worse performance of \acrshort{mppi} is most likely caused by the finite number of trajectories it can sample for each control evaluation. In contrast, gradient-based optimization lowers the cost iteratively and reaches a deeper minimum.

\begin{table} [h]
\centering
\label{tb:kpi}
\caption{Comparison of required time to damp the system and mean energy of the damped system.}
\begin{tabular}{c|ccc}
	Method & 80\% Energy & 90\% Energy & Damped\\
     &  Reduction &  Reduction & Mean \\
	\hline
	No Action & - & - & $5.2298\,\unit{J}$ \\
	MPPI & $323\,\unit{\ms}$ & $461\,\unit{\ms}$ & $1.0383\,\unit{J}$ \\ 
	Optimized & $120\,\unit{\ms}$ & $271\,\unit{\ms}$ & $0.8307 \,\unit{J}$ \\
	\hline
\end{tabular}
\end{table}

The computational efficiency for direct optimization with the differentiable simulator is also better than \acrshort{mppi}. Running both simulations on the same compute cluster with a Nvidia H200 GPU and 8 CPU cores, \acrshort{mppi} required roughly $9.5\,\unit{\hour}$ when parallelizing all trajectories. Our method using optimization took about $17\,\unit{\min}$. However, it should be noted that \acrshort{mppi} is a closed-loop control strategy, while our method is open-loop. Given the computational times we measured, we estimate that running optimization for this problem in closed-loop would take roughly $10000\,\unit{\s}$ or $2.8\,\unit{\hour}$ which would still be more than a $3\times$ speedup.

\section{Conclusion}
In this work, we have presented an adapted, differentiable \acrshort{mpm} that shows improved energy conservation at low spatial resolution, and we used it to optimize a control sequence for active damping. Our simulation results highlight that direct optimization achieves promising results in active vibration damping of a two-dimensional hyperelastic rope. Compared to \acrshort{mppi} optimization, achieves damping of the vibration around $2 \times$ faster and afterward keeps the kinetic energy in the rope $20\,\unit{\percent}$ lower. It further achieves these results while taking only about $3\,\unit{\percent}$ of the computation time the \acrshort{mppi} controller requires.
Future work will focus on improving the computational efficiency and demonstrating control based on trajectory optimization with the differentiable simulator on a real system.

\bibliographystyle{IEEEtran}
\bibliography{FullLibraryBibTex}

\end{document}